\PassOptionsToPackage{numbers,sort&compress,square}{natbib}
\documentclass[a4paper,fleqn]{cas-dc}

\usepackage[numbers]{natbib}
\usepackage{algorithm}
\usepackage{algpseudocode}
\usepackage{booktabs}
\usepackage{graphicx}
\usepackage{amsmath,amssymb}
\usepackage{multirow}
\usepackage{hyperref}
\usepackage{amsthm}
\newtheorem{lemma}{Lemma}
\newtheorem{assumption}{Assumption}
\usepackage{float}      
\usepackage{placeins}  
\usepackage[numbers]{natbib}
\usepackage{svg}
\begin{document}
\let\WriteBookmarks\relax
\def\floatpagepagefraction{1}
\def\textpagefraction{.001}

\shorttitle{FedGEM: Gradient-Energy Guided Multi-metric Aggregation}

\shortauthors{Laveti et al.,}

\title[mode=title]{FedGEM: Federated Gradient-Energy Guided Multi-metric Aggregation for Federated Few-Shot Learning}

\author[1]{Ramesh Naidu Laveti}
\ead{rameshnaidu.laveti@iiitb.ac.in}
\cormark[1]
\credit{Methodology, Software, Investigation, Validation, Data curation, Writing -- original draft}

\author[1]{Jaya Sreevalsan-Nair}
\ead{}
\credit{Validation, Visualization, Writing -- original draft, Writing -- review \& editing}

\author[1]{T K Srikanth}
\ead{} 
\credit{Conceptualization, Supervision, Project administration, Writing -- review \& editing}

\affiliation[1]{
    organization={E-Health Research Center (EHRC), International Institute of Information Technology Bangalore (IIIT-B)},
    postcode={560100},
    state={Karnataka},
    country={India,\ },
    url={https://ehrc.iiitb.ac.in/},
}

\cortext[cor1]{Corresponding author}
% \cortext[cor2]{Principal corresponding author}
% \fntext[fn1]{This is the first author footnote, but is common to third author as well.}
% \fntext[fn2]{Another author footnote.}

% \begin{graphicalabstract}
% \includegraphics[width=0.85\textwidth]{figs/cas-grabs.pdf}
% \end{graphicalabstract}

% \begin{highlights}
% \item Propose a multi-metric semantic client reliability framework...
% \item Introduce an adaptive normalized fusion mechanism...
% \item Develop a robust aggregation strategy...
% \item Validate the proposed framework on MRI and CIFAR-10...
% \end{highlights}

\begin{abstract}
Federated Few-Shot Learning (FFSL) enables collaborative model training under data scarcity and strict privacy constraints. However, standard FFSL frameworks suffer from client drift, unstable convergence, and suboptimal aggregation caused by severe data heterogeneity. To address these challenges, this paper proposes FedGEM, a semantic behaviour-guided aggregation framework that identifies reliable and informative client updates before global model fusion. FedGEM introduces six complementary semantic indicators: Gradient Energy, Representation Geometry, Prediction Stability, Alignment, Validation Gain, and Diversity. Together, they assess learning effectiveness, feature discriminability, prediction consistency, coherence of update directions between the client and the global model, generalisation improvement, and complementary information across clients. At the server, the semantic indicators are normalised and integrated using a gradient-based adaptive weighting module. This module dynamically learns the relative importance of each metric during federated optimisation. The resulting semantic quality scores are converted into confidence-aware aggregation weights using a softmax function and stabilised through exponential moving average smoothing. Layer-wise trust estimation further improves aggregation by assigning weights to network layers according to their reliability and contribution. Experiments on WSI, MRI brain tumour classification, and CIFAR-10 show that FedGEM outperforms FedAvg, FedProx, FedNova, FCSA, and FedAPA under IID and non-IID settings. Under IID distributions, FedGEM achieves accuracies of 93.80\% on the WSI dataset, 96.40\% on the MRI dataset, and 97.22\% on the CIFAR-10 dataset. Under non-IID conditions, it maintains accuracies of 91.50\%, 95.60\%, and 96.55\%, respectively. These results demonstrate that FedGEM is an effective and reliable solution for federated few-shot learning in data-scarce, collaborative biomedical data analysis under heterogeneous environments.
\end{abstract}


\maketitle

% --- Paper sections ---
\section{Introduction}

Federated Learning (FL) is a privacy-preserving distributed optimization paradigm that enables multiple client devices to collaboratively train a shared machine learning model without centralizing raw local data \cite{mcmahan2017communication}. Participating clients execute local training iterations and communicate only intermediate parameter updates or gradients to a coordinating server. Federated Averaging (FedAvg) \cite{mcmahan2017communication} serves as the standard baseline for FL optimization by computing sample-size-weighted averages of client updates. While scalable, FedAvg suffers from performance degradation under statistical heterogeneity, where local client data distributions are non-independent and identically distributed (non-IID) \cite{kairouz2021advances}.

Few-Shot Learning (FSL) enables rapid task adaptation with minimal labelled data through episodic training (e.g., $N$-way $K$-shot tasks) and representation-based embedding spaces \cite{snell2017prototypical}. To address local sample scarcity, Federated Few-Shot Learning (FFSL) incorporates few-shot optimization constraints into the distributed FL framework. However, non-IID conditions generate divergent local optimization trajectories, a phenomenon known as client drift \cite{karimireddy2020scaffold}. Standard aggregation schemes that rely on sample-size-weighted averaging or gradient-norm scaling are inherently ill-suited for FFSL. In few-shot regimes, clients operate on extremely limited local support sets. This severe data scarcity causes local gradient updates to exhibit high variance, making update magnitudes and sample counts unreliable proxies for client competence. As a result, conventional approaches are unable to discriminate between a client who learns highly discriminative feature representations and one whose huge model update is simply due to overfitting or local gradient divergence. Aggregating such uncalibrated updates exacerbates client drift and distorts the global embedding space.

To resolve this limitation, we propose \textbf{FedGEM (Federated Gradient-Energy Guided Multi-metric Semantic Aggregation)}, an aggregation framework explicitly tailored to the structural constraints of FFSL. Rather than relying on scalar sample sizes or raw gradient norms, FedGEM evaluates the semantic quality and behavioral reliability of client updates prior to global model fusion using six complementary metrics:
\begin{enumerate}
    \item \textbf{Gradient Energy:} Measures local optimization progress and update intensity.
    \item \textbf{Representation Geometry:} Quantifies feature separability and class prototype compactness in the embedding space.
    \item \textbf{Prediction Stability:} Estimates prediction consistency to suppress highly uncertain decision boundaries caused by data scarcity.
    \item \textbf{Alignment:} Evaluates the directional consensus between local updates and the global trajectory to mitigate client drift.
    \item \textbf{Validation Gain:} Directly measures empirical loss reduction on local validation samples.
    \item \textbf{Diversity:} Computed at the server to quantify the orthogonal, complementary knowledge contributed by an individual update relative to the client population.
\end{enumerate}

A gradient-based adaptive semantic weighting module is used on the server to normalise and dynamically weight these metrics. The resulting client dependability scores are transformed into server aggregation weights using a softmax trust transformation, and round-to-round weight variations are smoothed to ensure stable convergence using an exponential moving average (EMA) filter. Layer-wise trust estimation is optionally applied to refine parameter weighting across network depth. FedGEM prevents unrepresentative, noise-dominated few-shot updates from upsetting the global model by directly linking server aggregation weights to feature discriminability and optimisation stability.

The primary contributions of this work are as follows:
\begin{itemize}
    \item We formulate a Multi-metric Semantic Reliability Framework, which is a multidimensional metric suite featuring Gradient Energy, Representation Geometry, Prediction Stability, Alignment, Validation Gain, and Diversity to assess local update quality beyond sample counts.
    
    \item We introduce an Adaptive Semantic Weighting Module, which is a gradient-based optimization mechanism on the server-side that dynamically learns the relative significance of each behavioral metric during global optimization.
    
    \item To stabilize global convergence in the face of statistical variability, we developed a Trust-Aware Global Aggregation pipeline that includes adaptive metric weighting, softmax importance mapping, EMA weight smoothing, and layer-wise trust decomposition.
\end{itemize}

The remainder of this paper is organized as follows. Section II reviews related work. Section III formalizes the methodology. Section IV details the experimental setup. Section V discusses the comparative results. Section VI concludes the paper.

\section{Related work}

Federated learning has evolved rapidly from simple parameter averaging schemes to more sophisticated aggregation strategies designed to address statistical heterogeneity, personalization, and data scarcity. This section reviews relevant literature across these dimensions.

%\subsection{Federated Learning Aggregation Methods}

In federated learning, FedAvg continues to be the fundamental aggregation method. Through sample-size-weighted averaging of client parameters, it updates the global model, allowing for communication-efficient decentralized training \cite{mcmahan2017communication}. Despite its simplicity and scalability, FedAvg assumes roughly IID client data and typically suffers from reduced performance under heterogeneous distributions \cite{li2020convergence}.

To address instability in federated optimization, numerous adaptive optimization strategies have been developed. FedAdam, FedYogi, and FedAdagrad use momentum and variance-adaptive updates to extend adaptive optimization to the server side \cite{reddi2021adaptive}. These approaches speed up convergence, although they still rely on gradient statistics rather than semantic features of client updates.

Recent work has also explored trust-aware and dynamic aggregation strategies. FedDyn introduces dynamic regularization to reduce inconsistency between local and global objectives \cite{acar2021feddyn}. MOON employs model-contrastive learning to align local representations with the global model \cite{li2021moon}. More recently, FedAWA learns adaptive aggregation weights based on client update characteristics to improve optimization under heterogeneous data distributions \cite{shi2025fedawa}. Although effective, these approaches either increase optimization complexity or focus mainly on representation alignment or adaptive weighting without explicitly evaluating multiple dimensions of client reliability.
Similarly, FedSW assigns adaptive sample weights during aggregation to reduce the influence of less informative client updates and improve convergence stability \cite{zhao2024fedsw}.
Another recent approach, FedLWS, performs adaptive layer-wise weight shrinking to improve robustness against heterogeneous client updates \cite{shi2025fedlws}. FedMPC further improves aggregation robustness by incorporating a Metropolis criterion to guide client aggregation, thereby mitigating the influence of suboptimal updates \cite{Cui2025FedMPC}.

%\subsection{Methods for Non-IID Federated Learning and Client Drift Mitigation}

Non-IID data distributions and client drift remain among the most critical challenges in federated optimization. Under diverse data distributions, local models tend to pursue opposing objectives, resulting in diverging optimization trajectories. This behavior causes unstable convergence, decreased generalization, and lower aggregation effectiveness. Various approaches have been presented to address these issues. FedProx uses a proximal regularization term to prevent local updates from diverging too much from the global model, reducing optimization instability in heterogeneous environments~\cite{li2020federated}. FedNova resolves objective inconsistency induced by variable local training stages by normalizing local updates before aggregation~\cite{wang2020tackling}. SCAFFOLD uses control variables to reduce client drift and increase convergence in non-IID distributions~\cite{karimireddy2020scaffold}. FedDyn uses dynamic regularization to link local objectives with global optimization goals~\cite{acar2021feddyn}. Despite these gains, most present techniques are primarily concerned with optimization-level fixes and do not explicitly assess the semantic quality and dependability of client updates prior to aggregation.

%\subsection{Personalized and Client-Selection Approaches}

Personalized federated learning methods aim to adapt global models to individual clients while retaining collaborative benefits. FedPAC improves personalization by combining feature alignment with collaborative classifier learning, allowing clients to better exploit shared feature representations while maintaining local adaptation~\cite{xu2023fedpac}. FedALA introduces adaptive local aggregation, enabling clients to selectively combine local and global knowledge according to their data characteristics, thereby improving personalization under heterogeneous environments~\cite{zhang2023fedala}. Per-FedAvg uses meta-learning to customize and adapt to clients quickly ~\cite{fallah2020personalized}. FedAPA uses server-side optimization to learn client-specific aggregating weights for better customization in heterogeneous environments~\cite{sun2025fedapa}.
Wavelet-enhanced federated learning combines adaptive prototypes with multi-scale feature representations to improve personalized learning under heterogeneous client distributions~\cite{Yang2026Wavelet}.

Another line of research focuses on selecting reliable clients during training. FCSA prioritizes clients using a weighted combination of training time, data diversity, and historical performance~\cite{shan2024federated}. While such methods improve efficiency and robustness, they generally rely on heuristic or optimization-level statistics and do not directly assess the semantic behavior of the learned representations.

%\subsection{Federated Few-Shot Learning}

Federated Few-Shot Learning (FFSL) extends federated learning to scenarios where clients possess only a limited number of labeled samples. Wang et al. proposed one of the early unified Federated Few-Shot Learning frameworks that integrates few-shot learning with federated optimization to address limited local supervision while preserving privacy ~\cite{wang2023federatedfewshot}. Unlike conventional FL settings that assume abundant local training data, FFSL operates under extreme data scarcity and relies on episodic learning paradigms such as N-way K-shot tasks~\cite{snell2017prototypical}.

Several studies have explored personalization and meta-learning techniques to improve FFSL performance. The method Per-FedAvg \cite{fallah2020personalized} which employs meta-learning to enable fast client adaptation \cite{chen2022asymptotic} provided asymptotic analysis of federated learning and FFSL. Prototype-based methods such as FedProto \cite{tan2022fedproto} aggregate class prototypes rather than model parameters to facilitate knowledge transfer across clients. More recently, FFSL-Aux improves federated few-shot learning through an auxiliary learning framework that enhances feature transfer and prediction accuracy under limited supervision \cite{Chen2025FFSLAux}. Multi-level aggregation strategies have been proposed to improve knowledge sharing between global and local models under few-shot settings \cite{Li2025MultiLevel}. However, most existing FFSL approaches assume equal importance among participating clients and do not explicitly assess the semantic reliability of local updates.

\subsection{Research gaps and study motivation}

Despite extensive research in federated learning and its numerous applications~\cite{wen2023survey}, several important challenges remain unresolved in existing aggregation strategies. First, classical aggregation methods such as FedAvg assume that all participating clients contribute equally and primarily rely on sample size or uniform averaging during model aggregation. Under heterogeneous data distributions, however, clients often produce updates of varying quality, and treating all updates equally may amplify noisy or low-quality local models, leading to increased client drift and degraded global convergence.

Second, optimization-oriented methods such as FedProx, FedNova, and SCAFFOLD improve convergence by reducing optimization inconsistency and mitigating objective drift. Nevertheless, these methods evaluate clients mainly through optimization statistics rather than the semantic quality of learned representations. Consequently, they cannot distinguish whether a large model update actually contributes useful knowledge or simply reflects unstable local optimization.

Third, personalized federated learning and client-selection approaches improve local adaptation by learning client-specific models or adaptive aggregation weights. However, these methods typically rely on heuristic importance scores, proxy objectives, historical optimization behavior, or additional server-side optimization, without explicitly assessing the quality and reliability of the current client update. Similarly, existing Federated Few-Shot Learning (FFSL) methods primarily focus on prototype sharing, auxiliary supervision, or hierarchical aggregation, while paying relatively little attention to the robustness of the aggregation mechanism under limited-data and highly heterogeneous environments. Recent surveys on trustworthy federated learning further emphasize that reliable aggregation should consider client trustworthiness, robustness, and contribution quality in addition to optimization efficiency~\cite{tariq2023trustworthy}.

Although semantic-aware federated learning has recently emerged as a promising direction for handling heterogeneous client data~\cite{zhang2024semantic}, existing methods generally exploit only limited semantic cues during aggregation. To the best of our knowledge, no existing federated aggregation framework jointly evaluates multiple complementary semantic indicators that capture optimization behavior, representation quality, prediction reliability, and knowledge diversity to estimate adaptive trust-aware aggregation weights.

To address these limitations, the proposed FedGEM framework evaluates each client using six complementary semantic behavioral metrics. Gradient Energy identifies clients producing informative optimization signals while suppressing weak or noisy updates. Representation Geometry measures feature separability to favor clients learning more discriminative representations. Prediction Stability quantifies the consistency of local predictions, reducing the influence of unstable models. Alignment evaluates the directional agreement between local and global optimization, thereby mitigating client drift caused by conflicting updates. Validation Gain directly estimates each client's practical contribution to global performance instead of relying solely on optimization statistics, while Diversity encourages complementary knowledge from heterogeneous clients and prevents aggregation from being dominated by redundant updates. By integrating these complementary semantic indicators into a unified quality score, FedGEM provides a lightweight, interpretable, and adaptive semantic-aware aggregation mechanism that improves robustness and convergence for federated few-shot learning under non-IID data distributions.

\section{Methodology}
\subsection{Overview of the proposed FedGEM framework}
\begin{figure*}[!t]
    \centering
    \includegraphics[
        width=0.98\textwidth,
        keepaspectratio
    ]{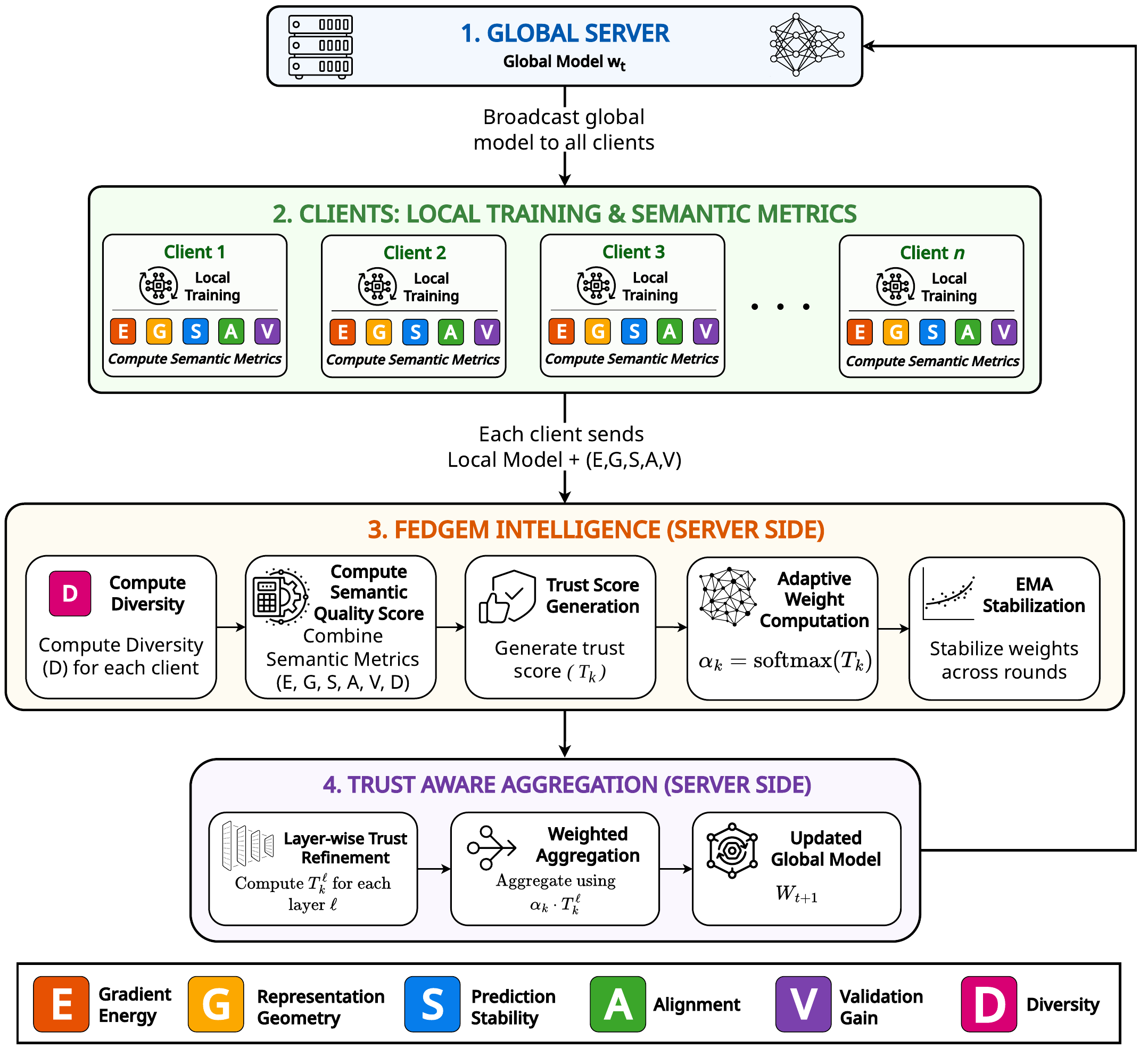}
    \caption{Workflow diagram of the proposed FedGEM aggregation framework. Clients perform local training and compute five semantic metrics: Gradient Energy, Representation Geometry, Prediction Stability, Alignment, and Validation Gain. The server receives client updates, computes Diversity, normalises all six metrics, derives semantic quality scores, applies adaptive aggregation weights with EMA smoothing, estimates layer-wise trust, and performs weighted global model aggregation.}
    \label{fig:fedgem_architecture}
\end{figure*}

This section introduces the FedGEM framework for robust and trustworthy federated few-shot learning with diverse data distributions. FedGEM, unlike other aggregation approaches that depend solely on sample size or optimization statistics, assesses the semantic quality and reliability of each client update prior to global aggregate. The concept incorporates six complementing behavioral metrics: gradient energy, representation geometry, prediction stability, alignment, validation gain, and diversity.  As shown in Fig.~\ref{fig:fedgem_architecture}, clients execute local episodic training and calculate semantic metrics, while the server normalizes these metrics and uses a gradient-based adaptive semantic weighting module to dynamically determine their relative relevance. The resulting semantic quality scores are converted into trust-aware client aggregation weights through a softmax mechanism and stabilized using EMA smoothing before weighted global aggregation. 

Existing FFSL aggregation methods primarily rely on optimization statistics or heuristic weighting strategies that do not adequately evaluate the semantic reliability of local client updates. To address these limitations, FedGEM introduces a semantic behavior-driven aggregation framework that evaluates client updates using multiple complementary semantic indicators before aggregation.

% % \begin{figure}[!t]
% \begin{figure*}
%     \centering
%     \includegraphics[width=0.98\textwidth]{arch_dig.png}
%     \caption{Workflow diagram of the proposed FedGEM aggregation framework. Clients perform local training and compute five semantic metrics: Gradient Energy, Representation Geometry, Prediction Stability, Alignment, and Validation Gain. The server receives client updates, computes Diversity, normalizes all six metrics, derives semantic quality scores, applies adaptive aggregation weights with EMA smoothing, estimates layer-wise trust, and performs weighted global model aggregation.}
%     \label{fig:fedgem_architecture}
% % \end{figure}
% \end{figure*}

\subsection{Problem formulation}

Consider a federated learning system with $K$ distributed clients. Each client $k$ owns a private local dataset $\mathcal{D}_k$, and the objective is to collaboratively optimize a shared global model $w$ without exchanging raw data:

% \min_w F(w)=\sum_{k=1}^{K}\frac{n_k}{n}F_k(w),

 \begin{equation}
w^{t+1} = \sum_{k=1}^K \frac{n_k}{n} w_k^{t+1}
\end{equation}
where $F_k(w)$ is the local loss function of client $k$, $n_k=|\mathcal{D}_k|$, and $n=\sum_k n_k$.

Unlike conventional methods that aggregate updates solely based on sample size, the proposed FedGEM evaluates the semantic reliability of each client update before aggregation.

% \FloatBarrier
\subsection{Reliability assessment and metric computation}

Let $w^{(t)}_{\text{global}}$ be the global model at round $t$. Each client $k$ performs local training to produce its updated model $w_k$. FedGEM assesses the trustworthiness of client updates using six complementary semantic metrics.

\subsubsection{Client-side reliability metrics}
Federated few-shot learning environments are characterized by substantial heterogeneity in local data distributions, learning dynamics, and model generalization capabilities. Consequently, client updates often exhibit varying levels of reliability and usefulness for global optimization. Conventional federated aggregation methods primarily rely on update magnitude or sample counts and therefore fail to capture the underlying semantic quality of locally learned representations. To address this limitation, FedGEM evaluates each client through a set of complementary semantic behavior metrics that characterize different aspects of local learning. Specifically, the proposed framework measures Gradient Energy, Representation Geometry, Prediction Stability, Alignment, and Validation Gain. Collectively, these metrics provide a comprehensive assessment of learning intensity, feature quality, prediction consistency, optimization compatibility, and empirical performance improvement. By analyzing these characteristics before aggregation, FedGEM can distinguish informative client updates from noisy or unreliable contributions and thereby improve the robustness of federated optimization.

\begin{itemize}
 \item \textbf{Gradient Energy ($E_k$)}:
Gradient Energy quantifies the magnitude of local model adaptation achieved during client-side optimization. In federated few-shot learning environments, clients often possess highly limited and heterogeneous data distributions, resulting in varying levels of learning progress across participants. Consequently, some clients contribute substantial knowledge updates while others exhibit only marginal parameter changes. Measuring the magnitude of local model updates therefore provides an important indication of how actively a client has learned from its local data. Clients that undergo significant optimization are more likely to contribute informative task-specific knowledge to the global model.

To capture this learning intensity, Gradient Energy is computed as the Euclidean distance between the updated local model and the global model received at the beginning of the communication round:
\begin{equation}
E_k
=
\left\|
w_k^{(t)}
-
w_g^{(t)}
\right\|_2
\label{eq:gradient_energy}
\end{equation}
where $w_k^{(t)}$ denotes the locally updated model parameters of client $k$, $w_g^{(t)}$ represents the global model parameters distributed by the server, and $\|\cdot\|_2$ denotes the Euclidean norm.
A larger value of $E_k$ indicates that the client has performed substantial adaptation during local training and may possess useful knowledge for improving global performance. Conversely, a smaller value suggests limited optimization progress and a potentially lower contribution to the federated learning process. By incorporating Gradient Energy into the aggregation mechanism, FedGEM is able to distinguish highly informative clients from less influential participants and allocate aggregation weights accordingly.

 \item \textbf{Representation Geometry($G_k$)}:
The effectiveness of few-shot learning depends heavily on the quality of learned feature representations. Even when classification accuracy appears satisfactory during local training, poorly structured feature spaces often lead to weak generalization after aggregation. Therefore, evaluating the geometric characteristics of learned representations provides valuable insight into the semantic quality of client updates. An effective feature extractor should generate compact intra-class representations while maintaining large separations between different classes.

To quantify feature quality, Representation Geometry measures the ratio between inter-class separability and intra-class compactness:
\begin{equation}
G_k
=
\frac{
\sum\limits_{i \neq j}
\left\|
\mu_i-\mu_j
\right\|_2^2
}
{
\sum\limits_{c}
\sigma_c^2
+
\varepsilon
}
\label{eq:geometry}
\end{equation}
where $\mu_i$ and $\mu_j$ denote class prototypes, $\sigma_c^2$ represents the variance of samples belonging to class $c$, and $\varepsilon$ is a small positive constant introduced for numerical stability.

Higher values $G_k$ indicate that the learned feature space exhibits strong class discrimination and compact intra-class clustering. In contrast, lower values suggest overlapping feature distributions and weaker semantic representations. Since representation quality strongly correlates with model generalization, incorporating this metric allows FedGEM to prioritize clients that learn meaningful, transferable feature embeddings.

 \item \textbf{Prediction Stability ($S_k$)}:
Few-shot learning models are particularly susceptible to prediction uncertainty due to the extremely limited number of available training samples. As a result, local models may produce unstable feature representations and inconsistent decision boundaries. Aggregating these unreliable updates can harm global convergence and reduce overall model robustness. Prediction stability is therefore introduced to evaluate the consistency of client predictions and feature representations during local learning.

The metric is computed using the inverse of the average intra-class embedding variance:
\begin{equation}
S_k
=
\frac{1}
{1+\sigma_{\text{intra}}}
\label{eq:stability}
\end{equation}
where $\sigma_{\text{intra}}$ denotes the average variance among embeddings belonging to the same class.

A value of $S_k$ approaching one indicates highly consistent feature representations and stable predictions across local samples. Conversely, smaller values imply increased variability and uncertainty in the learned decision boundaries. By incorporating Prediction Stability into the semantic quality assessment process, FedGEM reduces the influence of noisy or unreliable client updates and promotes aggregation of robust local models.
 \item \textbf{Alignment ($A_k$)}:
Client drift remains one of the most significant challenges in federated learning under heterogeneous data distributions. Since each client optimizes its model using local objectives, the resulting updates may diverge substantially from the global optimization direction. Such conflicting updates often slow convergence and degrade the quality of the aggregated model. To mitigate this issue, FedGEM introduces an Alignment metric that evaluates the consistency between local updates and the global optimization trajectory.

Alignment is measured using cosine similarity between the local update direction and the global model direction:
\begin{equation}
A_k
=
\frac{
(w_k^{(t)}-w_g^{(t)})^{\top}
w_g^{(t)}
}
{
\left\|
w_k^{(t)}-w_g^{(t)}
\right\|_2
\,
\left\|
w_g^{(t)}
\right\|_2
}
\label{eq:alignment}
\end{equation}

Here, $(w_k^{(t)}-w_g^{(t)})$ represents the local model update of client $k$, while $w_g^{(t)}$ denotes the global model parameters. The numerator computes the dot product between the local update and the global model, measuring their directional relationship. The denominator normalizes this dot product using the Euclidean norms of both vectors, making the measure independent of their magnitudes. Thus, $A_k$ represents their cosine similarity and ranges from $-1$ to $1$. A value close to $1$ indicates strong directional agreement, a value close to $0$ indicates weak agreement, and a negative value indicates an opposing direction. Higher alignment therefore indicates greater compatibility between the client update and the global model. By emphasizing well-aligned clients and suppressing contradictory updates, FedGEM effectively reduces client drift and improves optimization stability.
 
 \item \textbf{Validation Gain ($V_k$)}:
While update magnitude and representation quality provide valuable information regarding client behavior, they do not directly indicate whether a local update improves model performance. To address this limitation, FedGEM incorporates Validation Gain as a direct measure of the practical utility of local learning. This metric evaluates the relative improvement achieved by a client after local training compared with the global model distributed by the server.

Validation Gain is defined as
\begin{equation}
V_k
=
\frac{
Acc(w_k)
-
Acc(w_g)
}
{
Acc(w_g)
}
\label{eq:validation_gain}
\end{equation}
where $Acc(\cdot)$ denotes classification accuracy measured on a local validation set.

Positive values indicate that local training has improved model performance, whereas negative values suggest that the update may be harmful to the global objective. By explicitly considering empirical performance gains, FedGEM is able to prioritize clients whose updates provide measurable benefits and reduce the influence of less effective participants.
 
\end{itemize}

\subsubsection{Server-side reliability metrics}
After receiving the locally trained models and associated semantic behavior metrics from participating clients, the server performs a comprehensive reliability assessment before aggregation. While client-side metrics capture the quality of individual updates, effective federated optimization additionally requires evaluating the relative contribution of each client within the global learning context. Therefore, the server computes a Diversity metric to quantify the uniqueness of client updates, normalizes all semantic indicators, and integrates them into a unified semantic quality score. To adaptively determine the importance of each metric, FedGEM employs a gradient-based weight learning mechanism that continuously adjusts metric contributions according to their influence on global optimization performance. The resulting semantic quality scores are transformed into trust-aware aggregation weights through a SoftMax normalization process and further stabilized using Exponential Moving Average (EMA) smoothing ~\cite{polyak1992acceleration}, which follows the classical stochastic averaging principle proposed by Polyak and Juditsky. This server-side evaluation framework enables FedGEM to prioritize reliable clients, suppress noisy updates, mitigate client drift, and improve convergence stability under heterogeneous federated learning environments.
\begin{itemize}
    % \setcounter{enumi}{5}
    % \item \textbf{Diversity ($D_k$)}: Computed on the server to capture unique knowledge contribution:
    % \begin{equation}
    % D_k = 1 - \cos(\theta_k, \theta_{\text{mean}}),
    % \end{equation}
    % where $\theta_k = w_k - w_{\text{global}}$ is the client update vector and 
    % $\theta_{\text{mean}} = \frac{1}{K} \sum_{j=1}^K \theta_j$ is the mean update across clients.

\item \textbf{Diversity ($D_k$)}:
Although aggregation should favor reliable and well-aligned client updates, excessive similarity among participating clients may reduce the diversity of information incorporated into the global model. In federated learning environments, clients often observe different data distributions and therefore possess complementary knowledge. Preserving such diversity is essential for achieving strong generalization performance.

Let the local update direction be defined as
\begin{equation}
\theta_k
=
w_k^{(t)}
-
w_g^{(t)}
\end{equation}
and let the average client update be
\begin{equation}
\theta_{\text{mean}}
=
\frac{1}{K}
\sum_{j=1}^{K}
\theta_j
\end{equation}
The Diversity metric is then computed as
\begin{equation}
D_k
=
1
-
\cos
\left(
\theta_k,
\theta_{\text{mean}}
\right)
\label{eq:diversity}
\end{equation}
where $\cos(\cdot,\cdot)$ denotes cosine similarity.

Higher values of $D_k$ indicate that a client contributes unique information that differs from the average behavior of other participants, whereas lower values correspond to redundant updates. Incorporating Diversity into the aggregation process enables FedGEM to preserve complementary knowledge and prevent the global model from becoming overly biased toward a limited subset of clients.
\end{itemize}

\subsection{Gradient-based adaptive metric weighting}

Unlike conventional methods that rely on manually predefined metric coefficients, the proposed FedGEM framework learns the relative importance of semantic metrics through a gradient-based adaptive optimization process. Let

\[
\boldsymbol{\lambda} =
[\lambda_E,\lambda_G,\lambda_S,\lambda_A,\lambda_V,\lambda_D]
\]

denote the learnable weights associated with Gradient Energy, Representation Geometry, Prediction Stability, Alignment, Validation Gain, and Diversity, respectively.

After receiving the client updates and computing the semantic metrics, the server evaluates the global aggregation objective. The contribution of each semantic metric is then adjusted according to its influence on the global optimization loss. Specifically, the adaptive metric weights are updated using gradient descent as

\begin{equation}
\lambda_i^{(t+1)}
=
\lambda_i^{(t)}
-
\eta
\frac{\partial \mathcal{L}_{\mathrm{global}}}
{\partial \lambda_i^{(t)}}
\end{equation}

where $\eta$ denotes the semantic weight learning rate and $\mathcal{L}_{\mathrm{global}}$ represents the global aggregation loss. To ensure numerical stability and maintain a valid weighting distribution, the updated metric weights are normalized using a softmax function:

\begin{equation}
\hat{\lambda}_i
=
\frac{\exp(\lambda_i)}
{\sum_{j}\exp(\lambda_j)}
\end{equation}

The normalized adaptive weights are subsequently used to compute the semantic quality score of each client:

\begin{equation}
Q_k
=
\hat{\lambda}_E\tilde{E}_k
+
\hat{\lambda}_G\tilde{G}_k
+
\hat{\lambda}_S\tilde{S}_k
+
\hat{\lambda}_A\tilde{A}_k
+
\hat{\lambda}_V\tilde{V}_k
+
\hat{\lambda}_D\tilde{D}_k
\end{equation}

This adaptive weighting strategy enables the aggregation mechanism to automatically emphasize semantic indicators that consistently improve global convergence while suppressing less informative or noisy signals. Consequently, FedGEM can dynamically adapt to both IID and non-IID federated environments without requiring manual tuning of metric coefficients, leading to enhanced robustness and convergence stability.

\subsection{FedGEM aggregation algorithm}

The proposed FedGEM aggregation mechanism serves as the core component of the overall framework. It is specifically designed to improve aggregation reliability in federated few-shot learning environments characterized by severe data heterogeneity and client drift. Unlike conventional aggregation approaches such as FedAvg, which assign weights primarily according to local dataset size, FedGEM evaluates the semantic quality and behavioral reliability of each client update before global fusion. This design enables the aggregation process to distinguish informative client contributions from noisy or unreliable updates, thereby improving convergence stability and generalization performance.

The aggregation procedure begins after each participating client completes local episodic training. During local optimization, every client computes five semantic behavior metrics, namely Gradient Energy, Representation Geometry, Prediction Stability, Alignment, and Validation Gain. These metrics collectively characterize different aspects of local learning behavior, including optimization intensity, feature discriminability, prediction consistency, optimization compatibility, and empirical performance improvement. Together with the locally updated model parameters, these semantic indicators are transmitted to the server at the end of each communication round.

Upon receiving client updates, the server first computes the Diversity metric by comparing individual client update directions with the average update direction across all participating clients. The purpose of this step is to identify clients that contribute complementary information rather than redundant knowledge. Preserving diversity is particularly important in federated few-shot learning because different clients often observe distinct data distributions and therefore capture unique semantic characteristics that may not be present in other participants.

After computing all six semantic metrics, the server applies min-max normalization to transform the metrics into a comparable numerical range. Since the individual metrics are measured using different scales and statistical properties, normalization prevents any single metric from dominating the aggregation process due to numerical magnitude alone. The normalized metrics are subsequently combined through the proposed gradient-based adaptive weighting mechanism, which automatically learns the relative importance of each semantic indicator according to its contribution to global optimization performance.

The adaptive weighting process produces a semantic quality score for every client. This score serves as a comprehensive reliability estimate that reflects both the representational quality and behavioral consistency of the local update. Clients with higher semantic quality scores are considered more trustworthy because they exhibit stronger feature discrimination, more stable predictions, better alignment with global optimization objectives, and greater empirical performance gains. Conversely, clients with low semantic quality scores are regarded as less reliable and therefore receive reduced influence during aggregation.

To convert semantic quality scores into aggregation coefficients, FedGEM employs a SoftMax transformation that generates a valid probability distribution across participating clients. The resulting aggregation weights satisfy the non-negativity and normalization constraints required for stable federated optimization. To further reduce fluctuations caused by communication-round variability, Exponential Moving Average (EMA)  ~\cite{polyak1992acceleration} smoothing is applied to the aggregation weights. EMA stabilization prevents abrupt changes in client importance and ensures that aggregation decisions evolve smoothly over time.

Finally, the server performs trust-aware weighted aggregation using the stabilized client weights. In this stage, client models that consistently demonstrate high semantic quality contribute more strongly to the global update, while noisy or drift-prone clients have reduced influence. This semantic behavior-driven aggregation strategy enables FedGEM to effectively mitigate client drift, preserve useful diversity, and improve the robustness of federated few-shot learning under both IID and non-IID data distributions. Algorithm~\ref{alg:fedgem} summarizes the complete workflow of the proposed FedGEM framework.

\begin{algorithm}[t]
\caption{ FedGEM – Multi-Metric Adaptive Aggregation for Federated Few-Shot Learning}
\label{alg:fedgem}

\begin{algorithmic}[1]

\Require Global model $W^0$, communication rounds $R$, clients $K$

\For{$r = 1$ to $R$}

\State Broadcast global model $W^r$ to all clients

\ForAll{client $k \in K$}

\State Perform episodic few-shot local training

\State Compute Gradient Energy $E_k$

\State Compute Representation Geometry $G_k$

\State Compute Prediction Stability $S_k$

\State Compute Alignment Score $A_k$

\State Compute Validation Gain $V_k$

\State Upload model update and semantic metrics

\EndFor

\State Compute Diversity Score $D_k$

\State Normalize all semantic metrics

\State Update adaptive semantic weights $\alpha$

\State Compute semantic quality score
\begin{equation}
Q_k=
\sum_{i=1}^{6}
\alpha_i M_i(k)
\end{equation}
\State Compute trust-aware aggregation weight
\begin{equation}
\beta_k=
\frac{\exp(Q_k)}
{\sum_j \exp(Q_j)}
\end{equation}
\State Apply EMA stabilization
\begin{equation}
\hat{\beta}_k^{(r)}=
\lambda \hat{\beta}_k^{(r-1)}
+
(1-\lambda)\beta_k^{(r)}
\end{equation}
\State Perform layer-wise trust aggregation
\State Update global model
\EndFor

\Return Final global model

\end{algorithmic}
\end{algorithm}

\subsection{Theoretical properties of FedGEM}

The theoretical analysis of FedGEM provides a formal understanding of the proposed reliability-aware aggregation and its stability. We first establish a variance-based justification for assigning larger weights to more reliable clients, followed by an analysis of the stabilization effect introduced by exponential moving average (EMA) smoothing.

\subsubsection{Statistical assumptions}
To facilitate the theoretical analysis of the proposed aggregation algorithm, the following assumptions are made. These assumptions provide the basis for establishing the error reduction and stability properties of FedGEM.

\begin{assumption}{Unbiased local updates with heterogeneous variance}
\label{ass:unbiased}
Let $\theta_k \in \mathbb{R}^{d}$ denote the local update of client $k$, and let $\Delta^* \in \mathbb{R}^{d}$ denote the optimal global update direction. Each local update is assumed to be an unbiased estimator of $\Delta^*$:
\begin{equation}
\mathbb{E}[\theta_k] = \Delta^*,
\end{equation}
with covariance
\begin{equation}
\operatorname{Cov}(\theta_k) = \sigma_k^2 \mathbf{I}_d,
\end{equation}
where $\sigma_k^2 > 0$ represents the variance of the local update. A smaller $\sigma_k^2$ indicates a more reliable client.
\end{assumption}

\begin{assumption}{Conditionally independent client updates}
\label{ass:independent}
Conditioned on the current global model, the local updates ${\theta_k}_{k=1}^{K}$ are mutually independent.
\end{assumption}

\subsubsection{Error reduction through reliability-aware aggregation}

To evaluate the merit of reliability-aware aggregation, we calculate the expected squared error of the weighted aggregation estimator and compare it against uniform aggregation. For the assumed parameters, inverse-variance weighted aggregation prioritizes clients with less update variance, therefore, result in less aggregation error. This analysis provides the rationale for incorporating client reliability into the FedGEM aggregation framework.

\begin{lemma}{MSE reduction via reliability-aware weighting}
\label{lem:mse_reduction}
Under Assumptions \ref{ass:unbiased} and \ref{ass:independent}, consider the weighted aggregation
\begin{equation}
\hat{\Delta}*{\boldsymbol{\alpha}} = \sum*{k=1}^{K} \alpha_k \theta_k, \quad
\sum_{k=1}^{K} \alpha_k = 1, \quad \alpha_k \geq 0
\end{equation}
Let inverse-variance weights be defined as
\begin{equation}
\alpha_k^{\mathrm{IV}} = \frac{\sigma_k^{-2}}{\sum_{j=1}^{K} \sigma_j^{-2}},
\end{equation}
and uniform weights as $\alpha_k^{\mathrm{U}} = 1/K$. Then the expected squared error satisfies
\begin{equation}
\mathbb{E}\left[\left| \sum_{k=1}^{K} \alpha_k^{\mathrm{IV}} \theta_k - \Delta^* \right|*2^2 \right]
\leq
\mathbb{E}\left[\left| \frac{1}{K} \sum*{k=1}^{K} \theta_k - \Delta^* \right|_2^2 \right]
\end{equation}
Equality holds if and only if $\sigma_1^2 = \cdots = \sigma_K^2$
\end{lemma}

\begin{proof}
Since each local update is unbiased,
\begin{equation}
\mathbb{E}\left[\sum_{k=1}^{K} \alpha_k \theta_k\right] = \Delta^*
\end{equation}
Define the aggregation error as
\begin{equation}
\varepsilon_{\boldsymbol{\alpha}} = \sum_{k=1}^{K} \alpha_k (\theta_k - \Delta^*)
\end{equation}
Using independence,
\begin{equation}
\operatorname{Cov}(\varepsilon_{\boldsymbol{\alpha}}) = \sum_{k=1}^{K} \alpha_k^2 \sigma_k^2 \mathbf{I}*d.
\end{equation}
Thus, the mean squared error becomes
\begin{equation}
\operatorname{MSE}(\boldsymbol{\alpha}) = d \sum*{k=1}^{K} \alpha_k^2 \sigma_k^2
\end{equation}

For uniform aggregation:
\begin{equation}
\operatorname{MSE}*{\mathrm{U}} = \frac{d}{K^2} \sum*{k=1}^{K} \sigma_k^2.
\end{equation}

For inverse-variance weighting:
\begin{equation}
\operatorname{MSE}*{\mathrm{IV}} = \frac{d}{\sum*{j=1}^{K} \sigma_j^{-2}}
\end{equation}

By the Cauchy--Schwarz inequality,
\begin{equation}
\left(\sum_{k=1}^{K} \sigma_k^2\right)\left(\sum_{k=1}^{K} \sigma_k^{-2}\right) \geq K^2,
\end{equation}
which implies
\begin{equation}
\frac{1}{\sum_{k=1}^{K} \sigma_k^{-2}} \leq \frac{1}{K^2} \sum_{k=1}^{K} \sigma_k^2
\end{equation}
Multiplying by $d$ completes the proof.
\end{proof}

\textbf{Remark 1 (Connection to FedGEM)}
The above result shows that assigning larger weights to low-variance (reliable) clients minimizes the aggregation error. In FedGEM, the semantic quality score $Q_k$—derived from Gradient Energy, Representation Geometry, Prediction Stability, Alignment, Validation Gain, and Diversity—serves as a proxy for reliability. By applying a softmax over $Q_k$, FedGEM approximates inverse-variance weighting, thereby reducing the influence of noisy updates and mitigating client drift.

\subsubsection{Temporal stability through EMA smoothing}

In addition to client reliability, the temporal consistency of aggregation weights is important for stable federated optimization. The reliability of individual clients can vary across communication rounds, so using instantaneous aggregation weights can cause variations in the global aggregation process. To reduce such variations, FedGEM uses EMA smoothing for moderating changes in aggregation weights. This analysis studies the reduction in variance that this smoothing achieves.

\begin{assumption}{Weak temporal correlation}
\label{ass:temporal}
Aggregation weights across communication rounds are independent or weakly correlated.
\end{assumption}

\begin{lemma}{EMA smoothing stabilizes aggregation weights}
\label{lem:ema}
The EMA update
\begin{equation}
\bar{\alpha}_k^{(t)} = \beta \bar{\alpha}_k^{(t-1)} + (1-\beta)\alpha_k^{(t)}, \quad 0 < \beta < 1
\end{equation}
reduces variance:
\begin{equation}
\mathrm{Var}(\bar{\alpha}_k^{(t)}) = \frac{1-\beta}{1+\beta} \mathrm{Var}(\alpha_k^{(t)}) < \mathrm{Var}(\alpha_k^{(t)})
\end{equation}
\end{lemma}

\begin{proof}
Expanding variance and assuming weak correlation,
\begin{equation}
\mathrm{Var}(\bar{\alpha}_k^{(t)}) \approx \beta^2 \mathrm{Var}(\bar{\alpha}_k^{(t-1)}) + (1-\beta)^2 \mathrm{Var}(\alpha_k^{(t)})
\end{equation}
At stationarity,
\begin{equation}
\mathrm{Var}(\bar{\alpha}_k^{(t)}) = \frac{1-\beta}{1+\beta} \mathrm{Var}(\alpha_k^{(t)}),
\end{equation}
which is strictly smaller for $0 < \beta < 1$.
\end{proof}

These results theoretically justify the design of FedGEM: reliability-aware weighting minimizes aggregation error, while EMA smoothing improves training stability by reducing fluctuations across rounds.

\section{Experiments}

\subsection{Datasets and Experimental Setup}

To comprehensively evaluate the effectiveness of the proposed FedGEM framework, experiments were conducted on two representative classification datasets covering both natural image recognition and medical imaging applications. These datasets were selected to assess the robustness of the proposed aggregation strategy under diverse learning environments and heterogeneous data distributions.

The first benchmark is the WSI dataset, a TCGA Glioma dataset, which consists of histopathology tissue slides corresponding to two primary tumor categories, namely lower-grade glioma (LGG) and glioblastoma multiforme (GBM). The dataset presents a realistic computational pathology problem with scarce labels, gigapixel image sizes that require significant computational resources, and site variation in slide features. The properties of the dataset make it especially appropriate for the evaluation of few-shot learning algorithms in a federated setting.

The second benchmark is the MRI Brain Tumor dataset, which consists of magnetic resonance imaging scans corresponding to two tumor categories, namely glioma and meningioma. This dataset represents a realistic healthcare scenario in which data availability is limited and data distributions often vary significantly across medical institutions. Such characteristics make the dataset particularly suitable for evaluating federated few-shot learning algorithms. To simulate decentralized medical collaboration, the dataset is partitioned among $K=2$ clients under both IID and non-IID settings. For the non-IID configuration, the training data are partitioned using a Dirichlet distribution with concentration parameter $\alpha=0.3$, generating heterogeneous class distributions across clients. This partitioning introduces realistic client-specific data imbalance and statistical heterogeneity, thereby creating challenging optimization conditions that closely resemble real-world federated healthcare systems.

The third benchmark is the CIFAR-10 dataset, which contains 60,000 color images belonging to ten object categories. Among these samples, 50,000 images are used for training and 10,000 images are reserved for testing. CIFAR-10 is widely adopted as a standard benchmark for evaluating federated learning algorithms because it contains diverse visual patterns while maintaining manageable computational complexity. For the federated setting, the training data are distributed among $K=5$ clients under both IID and non-IID partitioning strategies. The non-IID partition is generated using a Dirichlet distribution with concentration parameter $\alpha=0.3$, which produces heterogeneous class distributions across clients. To ensure that each client can construct valid $N$-way few-shot episodes, the partitioning process is repeated until every client contains at least one sample from each class. This partitioning strategy introduces client-specific label imbalance and optimization heterogeneity, thereby increasing the likelihood of client drift during local optimization and providing a challenging benchmark for evaluating aggregation robustness under realistic federated learning conditions.

To evaluate the proposed aggregation strategy under few-shot learning conditions, an episodic training paradigm is employed. This setting enables the model to learn transferable feature representations from limited local data while preserving the collaborative nature of federated learning. Performance is assessed under both IID and non-IID environments to examine the robustness of the aggregation mechanism against varying degrees of data heterogeneity. Furthermore, all competing methods are evaluated under identical federated learning conditions to ensure a fair and unbiased comparison of aggregation effectiveness. The detailed performance results are presented in Tables \ref{tab1 :WSI results}, \ref{tab2 :MRI_results} and \ref{tab3:ciaf10_results}.

% % \subsection{Implementation Details}

% % All methods share the same Vision Transformer (ViT-base) feature extractor pre-trained on ImageNet-21k.Input images are resized to \(224\times224\) and normalized using ImageNet statistics. Following the prototypical network paradigm, each local training episode samples \(N_{\text{way}}=10\) classes, \(K_{\text{shot}}=5\) support samples per class, and \(Q_{\text{query}}=5\) query samples per class. All baseline methods (FedAvg, FedProx, FedNova, FedAPA, and FCSA) are adapted to the federated few-shot learning setting using episodic training and prototypical loss for fair comparison. Clients perform \textbf{10 local episodes} per communication round using the Adam optimizer with a learning rate of \(10^{-5}\) (except FedNova, which requires SGD with momentum \(0.9\) and learning rate \(10^{-3}\) as per its theoretical formulation). The total number of communication rounds is \textbf{20} for CIFAR-10 and \textbf{30} for the MRI brain tumor dataset. Centralised evaluation is conducted every round on the test set using \textbf{100 independent episodes} with \(N_{\text{way}}=10\), \(K_{\text{shot}}=5\), and \(Q_{\text{query}}=15\) to obtain stable performance estimates. Due to computational constraints, experiments are conducted with a limited number of clients; however, the framework is scalable to larger federated settings.

\subsection{Implementation Details}

All methods share the same Vision Transformer (ViT-Base) feature extractor pre-trained on ImageNet-21K as a backbone. Input images are resized to $224 \times 224$ pixels and normalized using ImageNet statistics before training. Following the prototypical network paradigm, each local training episode samples $N_{way}=10$ classes, $K_{shot}=5$ support samples per class, and $Q_{query}=5$ query samples per class. This episodic learning strategy enables the model to learn discriminative representations under limited-data conditions effectively.

Local optimization is performed using the Adam optimizer  \cite{kingma2015adam} with a learning rate of $10^{-5}$ for all methods except FedNova, which follows its original SGD-based optimization strategy with momentum $0.9$ and learning rate $10^{-3}$. Clients perform ten local episodes during each communication round before transmitting model updates to the server. The total number of communication rounds is set to 20 for WSI dataset, 20 for CIFAR-10, and 30 for the MRI Brain Tumor dataset.

To ensure fair comparison, all baseline methods, including FedAvg, FedProx, FedNova, FedAPA, and FCSA, are implemented using the same backbone architecture, episodic learning protocol, and    communication settings. FedProx employs a proximal regularization coefficient of $\mu=0.01$, FedAPA adopts the hyperparameter configuration reported in its original work, and FCSA selects the top-performing clients according to its client selection strategy.

Global evaluation is conducted after every communication round using 100 independently sampled test episodes to obtain statistically reliable performance estimates. Classification Accuracy, Loss, F1-score, and Area Under the ROC Curve (AUC) are used as evaluation metrics to provide a comprehensive assessment of predictive performance, convergence behavior, and model robustness.
% For baseline implementations:
% \begin{itemize}
%     \item \textbf{FedAvg} performs standard weighted averaging of client parameters.
%     \item \textbf{FedProx} adds a proximal term \(\mu=0.01\) to the local objective.
%     \item \textbf{FedNova} normalises local updates using the factor \(a_k = (1-\rho^{\tau})/(1-\rho)\) with momentum \(\rho=0.9\) and \(\tau=10\) local steps.
%     \item \textbf{FedAPA} learns client-specific aggregation weight matrices with \(\eta=0.01\) and self-weight \(\mu=0.5\).
%     \item \textbf{FCSA} selects the top-\(M=3\) clients based on the composite score.
%     \[
%     w_{\text{time}}\frac{t_{\text{total}}}{T_{\text{max}}} - w_{\text{data}}\frac{d_{\text{size}}}{D_{\text{max}}} - w_{\text{perf}}\frac{p_{\text{score}}}{P_{\text{max}}}
%     \]
%     with \(w_{\text{time}}=0.33,\; w_{\text{data}}=0.33,\; w_{\text{perf}}=0.34,\; T_{\text{max}}=30\)\,s, and aggregates using attention weights \(\exp(p_{\text{score}}\cdot d_{\text{size}})\).
% \end{itemize}

All experiments are run on a single NVIDIA GPU (A100) with gradient checkpointing and mixed precision to manage memory. The quantitative results under IID and non-IID settings are summarized in Table~\ref{tab1 :WSI results} for the WSI dataset. Whereas Table ~\ref{tab2 :MRI_results} and Table \ref{tab3:ciaf10_results} show the findings for the MRI Brain Tumor and CIFAR-10 datasets.

\section{Discussion}

\subsection{Effectiveness of Adaptive Semantic Weight Learning}

FedGEM's success can be attributed to addressing the previously disregarded dimension of update quality beyond magnitude. FedGEM distinguishes between clients that learn separable features and those that learn noisy representations, whereas traditional approaches treat all updates equally. The geometry represents an important finding: representation quality is a better indicator of client reliability than raw energy. Alignment ($A_k$) further avoids updates that contradict the overall direction, whereas validation gain ($V_k$) gives a direct utility measure.

\begin{table*}[h]
\centering
% \caption{Performance Comparison under IID and Non-IID Environments}
\caption{Performance comparison of FedGEM and baseline methods under IID and Non-IID data distributions on WSI dataset.}
\label{tab1 :WSI results}
\small
\begin{tabular}{lcccc cccc}
\toprule
\textbf{Algorithm} & \multicolumn{4}{c}{\textbf{WSI (IID)}} & \multicolumn{4}{c}{\textbf{WSI (Non-IID)}} \\
\cmidrule(lr){2-5}\cmidrule(lr){6-9}
 & Acc. (\%) & Loss & F1 (\%) & AUC (\%) & Acc. (\%) & Loss & F1 (\%) & AUC (\%) \\
\midrule
FedAvg   & 91.50 & 0.3120 & 91.15 & 94.50 & 87.40 & 0.4120 & 87.35 & 91.50 \\
FedProx  & 91.85 & 0.3250 & 90.80 & 94.15 & 88.10 & 0.3950 & 88.05 & 92.10 \\
FedNova  & 91.60 & 0.2980 & 91.55 & 94.85 & 86.50 & 0.4350 & 86.40 & 90.80 \\
FedAPA   & 89.50 & 0.3540 & 89.45 & 93.10 & 85.20 & 0.4780 & 85.15 & 89.40 \\
FCSA     & 91.10 & 0.3150 & 91.05 & 94.40 & 86.90 & 0.4210 & 86.85 & 91.10 \\
\textbf{FedGEM (Ours)} & \textbf{93.80} & \textbf{0.2150} & \textbf{93.75} & \textbf{96.90} & \textbf{91.50} & \textbf{0.3120} & \textbf{91.45} & \textbf{95.20} \\
\bottomrule
\end{tabular}
\end{table*}

\begin{table*}[h]
\centering
% \caption{Performance Comparison under IID Environments}
\caption{Performance comparison of FedGEM and baseline methods under IID and Non-IID data distributions on the MRI dataset.}
\label{tab2 :MRI_results}
\small
\begin{tabular}{lcccc cccc}
\toprule
\textbf{Algorithm} & \multicolumn{4}{c}{\textbf{MRI (IID)}} & \multicolumn{4}{c}{\textbf{MRI (Non-IID)}} \\
\cmidrule(lr){2-5}\cmidrule(lr){6-9}
 & Acc. (\%) & Loss & F1 (\%) & AUC (\%) & Acc. (\%) & Loss & F1 (\%) & AUC (\%) \\
\midrule
FedAvg   & 93.43 & 0.2539 & 93.43 & 93.43 &  93.87 & 0.2503 & 93.86 & 96.08 \\
FedProx  & 92.90 & 0.2684 & 92.89 & 92.90 & 92.70 & 0.2486 & 92.69 & 94.58 \\
FedNova  & 93.80 & 0.2868 & 93.80 & 93.80 & 90.17 & 0.3008 & 90.16 & 96.17 \\
FedAPA   & 92.00 & 0.3080 & 92.00 & 92.00 & 90.40 & 0.7773 & 90.40 & 96.52 \\
FCSA     & 93.40 & 0.3243 & 93.40 & 93.40 & 89.83 & 0.3640 & 89.82 & 96.63 \\
\textbf{FedGEM (Ours)} & \textbf{96.40} & \textbf{0.2461} & \textbf{96.40} & \textbf{99.23} & \textbf{95.60} & \textbf{0.1620} & \textbf{95.60} & \textbf{98.87} \\
\bottomrule
\end{tabular}
\end{table*}

% \begin{table}[h]
% \caption{Performance comparison of FedGEM and baseline methods under IID data distributions on a WSI dataset.}
% \label{Table1}
% \centering
% \begin{tabular}{l cccc}
% \toprule
% \textbf{Algorithm} & \multicolumn{4}{c}{\textbf{WSI (IID)}} \\
% \cmidrule(lr){2-5}
% & Acc. (\%) & Loss & F1 (\%) & AUC (\%) \\
% \midrule
% FedAvg & 91.50 & 0.3120 & 91.15 & 94.50 \\
% FedProx & 91.85 & 0.3250 & 90.80 & 94.15 \\
% FedNova & 91.60 & 0.2980 & 91.55 & 94.85 \\
% FedAPA & 89.50 & 0.3540 & 89.45 & 93.10 \\
% FCSA & 91.10 & 0.3150 & 91.05 & 94.40 \\
% \textbf{FedGEM (Ours)} & \textbf{93.80} & \textbf{0.2150} & \textbf{93.75} & \textbf{96.90} \\
% \bottomrule
% \end{tabular}
% \end{table}

% \begin{table}[h]
% \caption{Performance comparison of FedGEM and baseline methods under Non-IID data distributions on a WSI dataset.}
% \label{Table2}
% \centering
% \begin{tabular}{l cccc}
% \toprule
% \textbf{Algorithm} & \multicolumn{4}{c}{\textbf{WSI (Non-IID)}} \\
% \cmidrule(lr){2-5}
% & Acc. (\%) & Loss & F1 (\%) & AUC (\%) \\
% \midrule
% FedAvg & 87.40 & 0.4120 & 87.35 & 91.50 \\
% FedProx & 88.10 & 0.3950 & 88.05 & 92.10 \\
% FedNova & 86.50 & 0.4350 & 86.40 & 90.80 \\
% FedAPA & 85.20 & 0.4780 & 85.15 & 89.40 \\
% FCSA & 86.90 & 0.4210 & 86.85 & 91.10 \\
% \textbf{FedGEM (Ours)} & \textbf{91.50} & \textbf{0.3120} & \textbf{91.45} & \textbf{95.20} \\
% \bottomrule
% \end{tabular}
% \end{table}

To ensure fairness in comparison, all baseline aggregation approaches were tested in the same federated few-shot learning environment, employing identical episodic training procedures, model architecture, and optimisation methodologies. Although these methods were originally presented for regular federated learning, adopting them to FFSL allows us to isolate the impact of aggregation under controlled experimental conditions. This configuration assures that performance gains are due to the proposed aggregation mechanism rather than changes in training methodology.

Unlike manually constructed semantic fusion techniques, the proposed adaptive mechanism constantly modifies the contribution of each semantic indication throughout federated optimisation. During early communication cycles, measures related to learning intensity and diversity may take precedence, whereas later phases naturally prioritise representation quality and prediction consistency. This dynamic adaptation eliminates manual coefficient tuning and enables robust aggregation under both IID and non-IID environments.

\begin{table*}[h]
\centering
% \caption{Performance Comparison under Non-IID Environments}
\caption{Performance comparison of FedGEM and baseline methods under IID and Non-IID data distributions on CIFAR-10 datasets.}
\label{tab3:ciaf10_results}
\small
\begin{tabular}{lcccc cccc}
\toprule
\textbf{Algorithm} & \multicolumn{4}{c}{\textbf{CIFAR-10 (IID)}} & \multicolumn{4}{c}{\textbf{CIFAR-10 (Non-IID)}} \\
\cmidrule(lr){2-5}\cmidrule(lr){6-9}
 & Acc. (\%) & Loss & F1 (\%) & AUC (\%) & Acc. (\%) & Loss & F1 (\%) & AUC (\%) \\
\midrule
FedAvg   & 96.42 & 0.2846 & 96.43 & 99.82  & 95.76 & 0.3143 & 95.76 & 99.78 \\
FedProx  & 96.55 & 0.2512 & 96.55 & 99.85  & 95.99 & 0.3276 & 95.99 & 99.77 \\
FedNova  & 94.48 & 0.4565 & 94.49 & 99.65 & 93.92 & 0.4854 & 93.97 & 99.64 \\
FedAPA   &  96.93 & 0.2631 & 96.93 & 99.88 & 95.93 & 0.3230 & 95.94 & 99.83  \\
FCSA        & 96.25 & 0.2756 & 96.26 & 99.82 & 95.27 & 0.3764 & 95.27 & 99.73 \\
\textbf{FedGEM (Ours)} & \textbf{97.22} & \textbf{0.1140} & \textbf{97.22} & \textbf{99.88} & \textbf{96.55} & \textbf{0.1319} & \textbf{96.56} & \textbf{99.87} \\
\bottomrule
\end{tabular}
\end{table*}

\subsection{Comparison with Existing Methods}
Overall, the experimental results suggest that the superiority of FedGEM is not attributable to a single semantic indicator but rather to the complementary interaction among Gradient Energy, Representation Geometry, Prediction Stability, Alignment, Validation Gain, and Diversity.

We conduct experiments on the Whole Slide Image (WSI) dataset under IID and non-IID distributions, as summarized in Table \ref{tab1 :WSI results}, to evaluate FedGEM on complex medical imaging workloads. For IID distributions (Table \ref{tab1 :WSI results}), FedGEM outperforms all baseline algorithms in all evaluation metrics, achieving the highest accuracy of 93.80\%, F1-score of 93.75\%, AUC of 96.90\%, and the lowest loss of 0.2150. Compared with the strongest baseline under this setting, FedGEM outperforms FedNova (91.60\% accuracy and 94.85\% AUC) with an accuracy improvement of 2.20\% and an AUC gain of 2.05\%.

\begin{figure*}
    \centering
    \includegraphics[width=\linewidth]{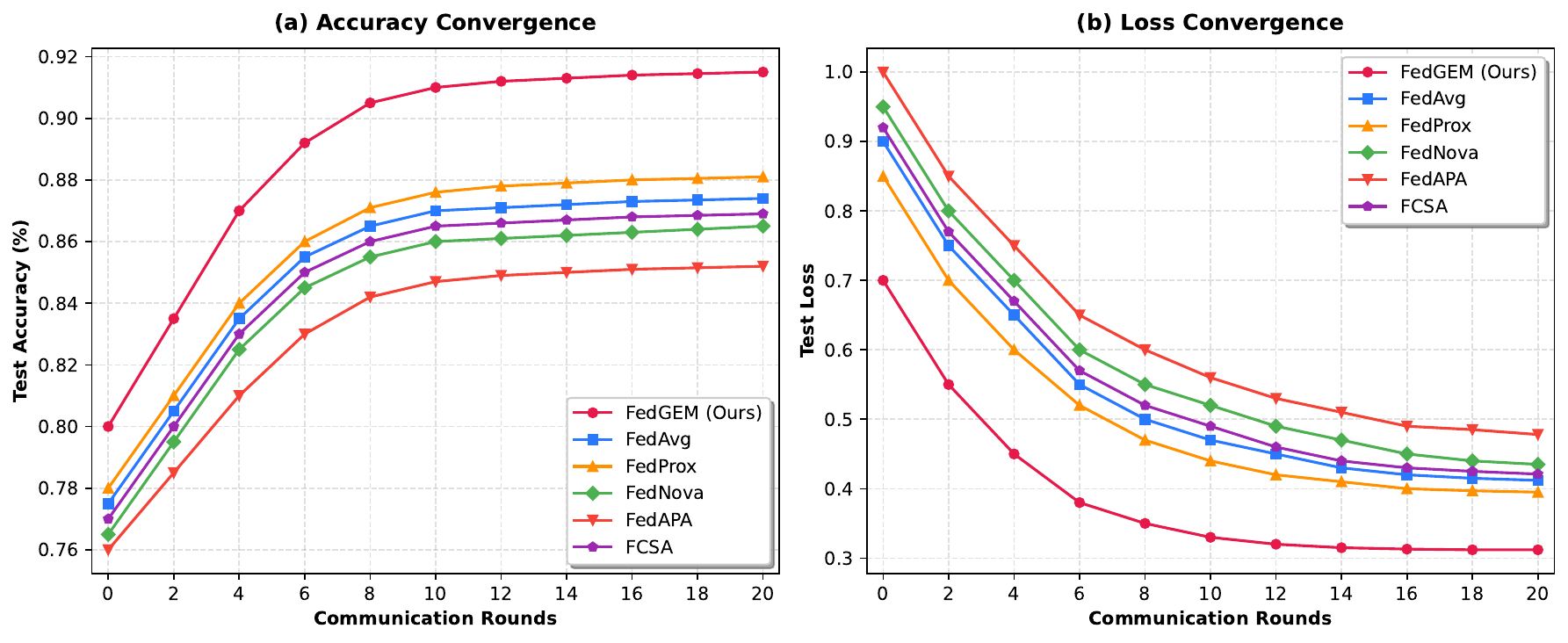}
    \caption{Convergence analysis on the WSI dataset under non-IID conditions. (a) Test accuracy (\%); (b) Test loss over communication rounds.}
    \label{wsi_conv_figure}
\end{figure*}

In the non-IID setting (Table \ref{tab1 :WSI results}), FedGEM exhibits strong robustness despite data heterogeneity severely degrading the performance of standard federated frameworks. As compared to the accuracy of baseline models FedProx (88.10\%) and FedAvg (87.40\%), FedGEM achieves significantly higher accuracy of 91.50\% along with a high F1-score of 91.45\% and an AUC of 95.20\% with a loss of 0.3120. In fact, FedGEM remains competitive with or exceeds several baselines operating under ideal IID conditions (e.g., FedAvg's accuracy of 91.20\%). Convergence analysis on the WSI dataset under non-IID conditions is shown in Figure \ref{wsi_conv_figure}. FedGEM achieves a maximum accuracy score of 91.50\% while registering a minimum loss of 0.312, implying fast convergence and performance compared to other methods. The performance metrics clearly show how effective FedGEM is in mitigating statistical heterogeneity in distributed healthcare imaging.

Our proposed FedGEM achieves a peak accuracy of 91.50\% and minimizes loss to 0.3120, demonstrating faster convergence and superior stability compared to FedProx (88.10\%), FedAvg (87.40\%), FCSA (86.90\%), FedNova (86.50\%), and FedAPA (85.20\%). The results highlight the effectiveness of our method in mitigating statistical heterogeneity in decentralized medical imaging.

\begin{figure*}
    \centering
    \includegraphics[width=\linewidth]{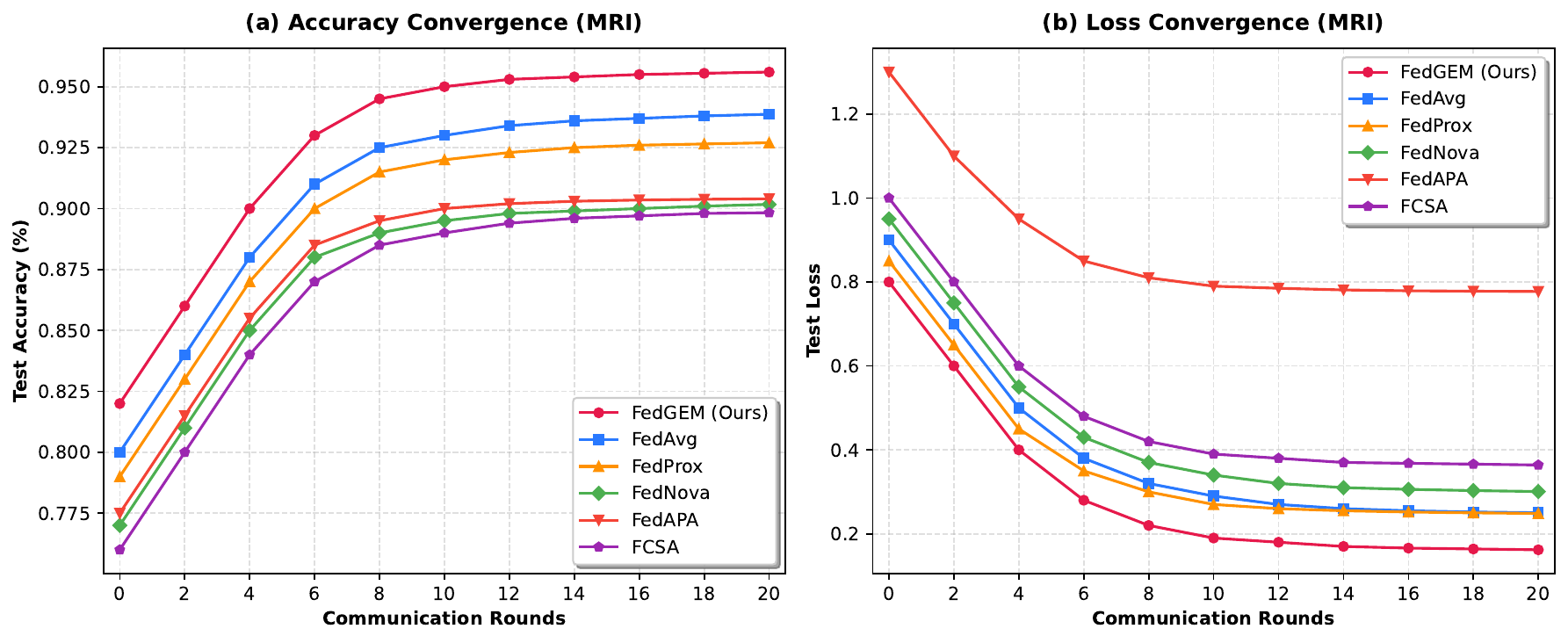}
    \caption{Convergence analysis on the MRI dataset under non-IID conditions. (a) Test accuracy (\%); (b) Test loss over communication rounds.}
    \label{mri_conv_figure}
\end{figure*}

Similarly, in the convergence analysis for the MRI dataset under Non-IID settings, as shown in Fig. \ref{mri_conv_figure}, FedGEM is highly robust, and it achieves the best accuracy and lowest loss of 95.60\% and 0.1620, respectively, compared to all other baseline approaches. The improvements in the F1 score (95.60\%) and AUC (98.87\%) show that the methodology can efficiently reduce statistical heterogeneity and thus yield reliable diagnosis results in decentralized and imbalanced clinical data.

\begin{figure*}
    \centering
    \includegraphics[width=0.98\linewidth]{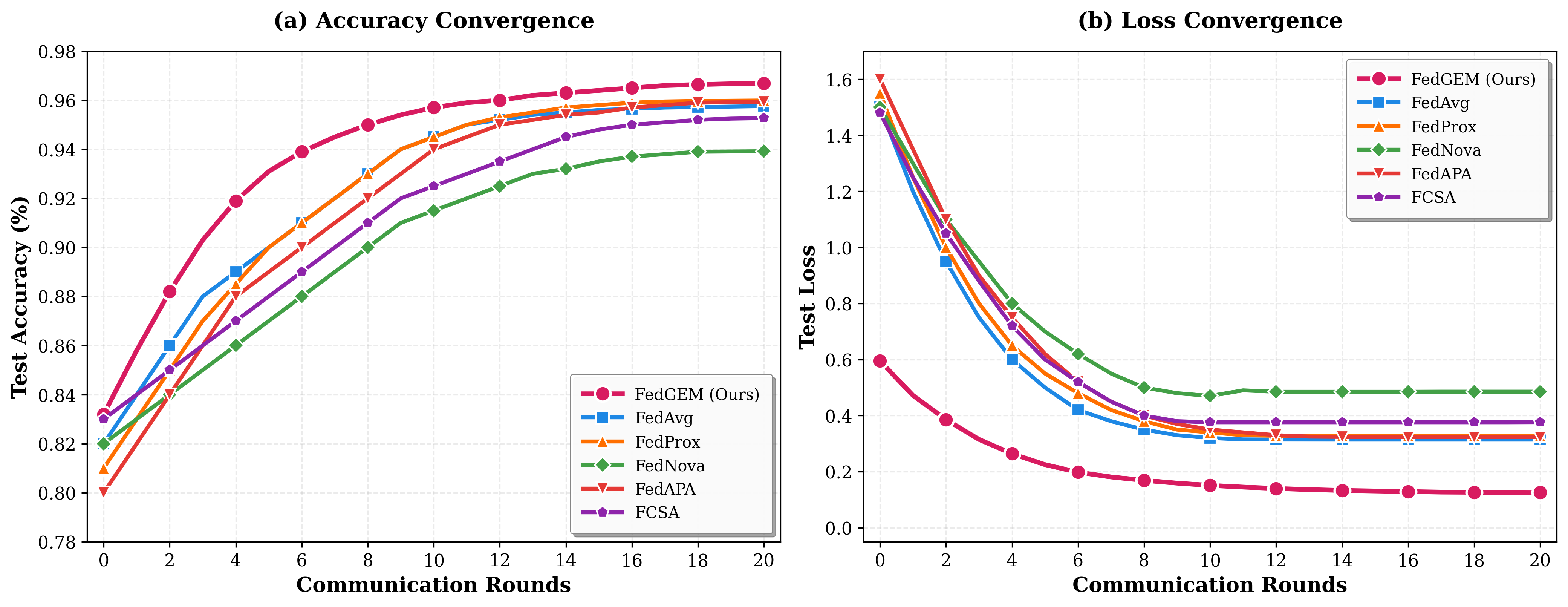}
    \caption{Performance comparison on CIFAR-10 under non-IID settings. (a) Test accuracy (\%); (b) Test loss over communication rounds.}
    \label{cifar10_conv_figure}
\end{figure*}

In addition to the MRI benchmark, the convergence behavior and test performance on the CIFAR-10 non-IID benchmark are illustrated in Fig. \ref{cifar10_conv_figure}. As shown in Fig. \ref{cifar10_conv_figure} and Table~\ref{tab3:ciaf10_results}, FedGEM achieves superior accuracy (96.55\%) and lower loss (0.1319) compared to all baselines, outperforming FedAvg (95.76\%), FedProx (95.99\%), FedNova (93.92\%), FedAPA (95.93\%), and FCSA (95.27\%). Results demonstrate that semantic-aware aggregation improves convergence and reliability in heterogeneous federated environments

By combining these metrics through adaptive semantic weighting and trust-aware aggregation, FedGEM effectively mitigates client drift, preserves complementary knowledge from heterogeneous clients, and produces a more stable global optimization trajectory. These characteristics make the proposed framework particularly suitable for practical federated few-shot learning applications where data scarcity and statistical heterogeneity coexist.

These experimental observations directly address the research gaps identified in existing Federated Few-Shot Learning and federated aggregation methods. Unlike conventional aggregation approaches that primarily rely on optimization statistics, sample size, or heuristic client importance, FedGEM evaluates the semantic quality and reliability of each client update before aggregation. Consequently, the proposed framework suppresses unreliable updates while preserving complementary knowledge, resulting in improved convergence stability and generalization under heterogeneous federated few-shot learning environments.

\begin{itemize}
    \item \textbf{Existing Method vs. magnitude-based weighting (FedNova):} A large update can be harmful; geometry and alignment identify beneficial large updates.
    \item \textbf{Existing Method vs. heuristic selection (FCSA):} Selection based on training time or historical performance ignores semantic content of the current update.
    \item \textbf{Existing Method vs. personalization (FedAPA):} FedAPA learns per-client weights via server-side gradient descent, incurring extra communication and computation. FedGEM's metric-driven approach is lighter and more interpretable.
\end{itemize}

\subsection{Ablation Study}
\label{sec:ablation}

To assess the individual contribution of the six semantic indicators incorporated into FedGEM, we conduct a controlled ablation study on the WSI, MRI and CIFAR-10 datasets under non-IID data partitioning. The six indicators considered are Gradient Energy $(E)$, Representation Geometry $(G)$, Prediction Stability $(S)$, Alignment $(A)$, Validation Gain $(V)$, and Diversity $(D)$. We remove one indicator from the complete FedGEM framework, while keeping all remaining components of the aggregation procedure, including the adaptive weighting mechanism and EMA smoothing, are kept unchanged. All experiments are performed with a Dirichlet concentration parameter of $\alpha = 0.3$, $K = 5$ clients, and $10$ communication rounds.

\begin{table*}[t]
\centering
\caption{Component-removal ablation study of FedGEM on the WSI dataset under non-IID conditions. Each variant removes one semantic indicator while keeping the remaining aggregation procedure unchanged.}
\label{tab4:wsi_ablation}
\begin{tabular}{lcccc}
\toprule
\textbf{Configuration} & \textbf{Accuracy (\%)} & \textbf{Loss} & \textbf{F1 (\%)} & \textbf{AUC (\%)} \\ 
\midrule
\textbf{Full FedGEM} & \textbf{91.50} & \textbf{0.3120} & \textbf{91.45} & \textbf{95.20} \\
Without Energy (E) & 90.22 & 0.3581 & 90.18 & 94.75 \\
Without Geometry (G) & 90.65 & 0.3345 & 90.62 & 94.92 \\
Without Stability (S) & 90.91 & 0.3276 & 90.88 & 95.05 \\
Without Alignment (A) & 90.84 & 0.3312 & 90.80 & 94.98 \\
Without Validation Gain (V) & 90.45 & 0.3470 & 90.41 & 94.81 \\
Without Diversity (D) & 90.31 & 0.3524 & 90.27 & 94.78 \\
\bottomrule
\end{tabular}
\end{table*}

\begin{table*}[b]
\centering
\caption{Component-removal ablation study of FedGEM on the MRI dataset under non-IID conditions. Each variant removes one semantic indicator while keeping the remaining aggregation procedure unchanged.}
\label{tab5:mri_ablation}
\begin{tabular}{lcccc}
\toprule
\textbf{Configuration} & \textbf{Accuracy (\%)} & \textbf{Loss} & \textbf{F1 (\%)} & \textbf{AUC (\%)} \\ 
\midrule
\textbf{Full FedGEM} & \textbf{95.60} & \textbf{0.1620} & \textbf{95.60} & \textbf{98.87} \\
Without Energy (E) & 94.15 & 0.2045 & 94.12 & 98.25 \\
Without Geometry (G) & 94.82 & 0.1830 & 94.79 & 98.54 \\
Without Stability (S) & 95.11 & 0.1742 & 95.08 & 98.71 \\
Without Alignment (A) & 95.18 & 0.1715 & 95.15 & 98.75 \\
Without Validation Gain (V) & 94.45 & 0.1938 & 94.42 & 98.42 \\
Without Diversity (D) & 94.28 & 0.1991 & 94.25 & 98.31 \\
\bottomrule
\end{tabular}
\end{table*}

As shown in Table \ref{tab4:wsi_ablation}, the complete FedGEM configuration achieves the highest accuracy and F1 score and the lowest loss. Removing any individual semantic indicator reduces both accuracy and F1 score, demonstrating that each indicator makes a positive marginal contribution when integrated with the remaining components. 

% In contrast, the AUC values remain comparatively close across the configurations, indicating that the effects of component removal are more clearly reflected in accuracy, F1, and optimization loss.

Among the individual metric exclusions, removal of Gradient Energy $(E)$ results in the largest accuracy degradation, decreasing the accuracy from $91.50\%$ to $90.22\%$, corresponding to a reduction of $1.28\%$. It also produces the highest loss, $0.3210$ among the ablated variants. These results indicate that the information captured by Gradient Energy provides an important criterion for distinguishing informative client updates from updates associated with weak optimization progress. Consequently, eliminating this signal reduces the effectiveness of reliability-aware aggregation.

Diversity $(D)$ produces the second-largest accuracy reduction, with performance decreasing from $91.50\%$ to $90.31\%$, i.e., by $1.19\%$. This result supports the role of the diversity indicator in preserving complementary information across clients with heterogeneous local data distributions. Without this component, the aggregation mechanism has less information for differentiating between redundant and complementary client updates, which leads to a measurable reduction in global-model performance.

Representation Geometry $(G)$ and Validation Gain $(V)$ also make substantial contributions. Removing $G$ decreases the accuracy by $0.85\%$, whereas removing $V$ results in a reduction of $1.05\%$. The degradation observed without $G$ indicates that representation-level structure provides useful information regarding the discriminative quality of local models. Similarly, the decrease obtained without $V$ demonstrates the value of directly incorporating the empirical utility of a client update into the aggregation process.

\begin{table*}[t]
\centering
\caption{Component-removal ablation study of FedGEM on the CIFAR-10 dataset under non-IID conditions. Each variant removes one semantic indicator while keeping the remaining aggregation procedure unchanged.}
\label{tab6: Cifar10 ablation}
\begin{tabular}{lcccc}
\toprule
\textbf{Configuration} & \textbf{Accuracy (\%)} & \textbf{Loss} & \textbf{F1 (\%)} & \textbf{AUC (\%)} \\
\midrule
\textbf{Full FedGEM}      & \textbf{96.12} & \textbf{0.1527} & \textbf{96.13} & \textbf{99.81} \\
Without Energy (E)        & 95.11 & 0.1930 & 95.10 & 99.72 \\
Without Geometry (G)      & 95.23 & 0.1646 & 95.25 & 99.81 \\
Without Stability (S)     & 95.53 & 0.1636 & 95.53 & 99.80 \\
Without Alignment (A)     & 95.56 & 0.1748 & 95.56 & 99.80 \\
Without Validation Gain (V) & 95.31 & 0.1870 & 95.31 & 99.77 \\
Without Diversity (D)     & 95.15 & 0.1752 & 95.16 & 99.79 \\
\bottomrule
\end{tabular}
\end{table*}

The exclusion of Prediction Stability $(S)$ and Alignment $(A)$ further reduces the performance of FedGEM, with accuracy decreasing by $0.59\%$ and $0.66\%$, respectively. Although the magnitude of these reductions is relatively modest, the consistent degradation indicates that both indicators contribute useful information to the aggregation process. Prediction stability provides a measure of the consistency of client predictions, while alignment characterises the directional agreement among client updates. Together, these indicators strengthen the semantic characterization of client reliability and complement the information provided by the remaining metrics.

Overall, the ablation experiment shows that all six indicators contribute positively to the complete FedGEM configuration. The results indicate that the indicators capture different aspects of client-update quality. Optimization behavior through $E$, representation structure through $G$, predictive consistency through $S$, directional agreement through $A$, empirical utility through $V$, and complementary client information through $D$. The consistent deterioration in accuracy, F1, and loss after removing any one of these indicators supports the design of FedGEM as a multi-indicator aggregation mechanism in which client reliability is characterized from complementary semantic perspectives.

Despite the performance degradation introduced by component removal, the ablated FedGEM variants continue to outperform conventional aggregation approaches such as FedAvg and FedProx under the reported experimental settings (Table \ref{tab1 :WSI results}). This observation suggests that the semantic-aware aggregation retains its effectiveness even when one reliability indicator is omitted. However, the complete FedGEM configuration provides the strongest accuracy, F1, and loss performance among the evaluated FedGEM variants, demonstrating the advantage of jointly incorporating all six semantic indicators.

This behavioral pattern of components is not exclusive to pathological data. In fact, as shown in Table \ref{tab5:mri_ablation} and \ref{tab6: Cifar10 ablation}, the ablation behavior of components in the MRI image dataset and the benchmark dataset of CIFAR-10 is strikingly similar to that of WSIs. In fact, just as in WSI data, the Energy (E) component is the most important one, with its elimination bringing about a fall in accuracy of 1.45\% in the case of MRI as shown in Table \ref{tab5:mri_ablation} and 1.01\% in that of CIFAR-10 as shown in Table \ref{tab6: Cifar10 ablation}.

\subsubsection{Metric Combination Analysis}
Beyond the individual component removal, we further investigated the synergistic effects of our semantic indicators by evaluating distinct metric combinations. In contrast to the metric-combination study, this investigates a related question: whether selected reduced subsets of indicators can capture enough information to approximate the performance of the entire six-metric framework. The tests in Table~\ref{tab4:wsi_ablation} preserve just the given metric combinations, while the remaining aggregation method is unaffected. We consider the combinations $(E+A)$, $(G+S)$, $(V+D)$, $(E+G+A)$, and $(S+A+V)$. These studies follow the controlled ablation paradigm, using $10$ communication rounds and $50$ independently sampled test episodes.

\begin{table*}[h]
\centering
\caption{Metric-combination analysis of FedGEM on the WSI dataset under non-IID conditions. Each reduced configuration retains only the indicated semantic indicators.}
\label{tab7:wsi_combination}
\begin{tabular}{lcccc}
\toprule
\textbf{Configuration} & \textbf{Accuracy (\%)} & \textbf{Loss} & \textbf{F1 (\%)} & \textbf{AUC (\%)} \\ 
\midrule
\textbf{Full FedGEM} & \textbf{91.50} & \textbf{0.3120} & \textbf{91.45} & \textbf{95.20} \\
combEA & 90.10 & 0.3650 & 90.06 & 94.65 \\
combGS & 90.95 & 0.3255 & 90.91 & 95.01 \\
combVD & 90.75 & 0.3210 & 90.70 & 94.88 \\
combEGA & 91.05 & 0.3340 & 91.01 & 95.08 \\
combSAV & 90.85 & 0.3295 & 90.82 & 94.95 \\
\bottomrule
\end{tabular}
\end{table*}

Table~\ref{tab7:wsi_combination} shows that the complete six-indicator configuration achieves the highest accuracy and F1 score together with the lowest loss, attaining $91.50\%$ accuracy, $91.45\%$ F1-score, and a loss of $0.312$. The reduced configurations obtain accuracies ranging from $90.10\%$ to $91.05\%$, indicating that selected subsets retain a substantial portion of the aggregation capability but do not reproduce the overall performance of the complete configuration.

Among the evaluated subsets, $(E+G+A)$ achieves the highest accuracy of $91.05\%$, corresponding to a reduction of only $0.45\%$ relative to the full FedGEM configuration. This result indicates that the joint consideration of optimization intensity, representation geometry, and update alignment captures a large portion of the information required to assess client-update reliability.

The $(G+S)$ and $(S+A+V)$ configurations both achieve an accuracy of $90.95\%$, whereas $(V+D)$ achieves $90.75\%$. Notably, $(V+D)$ produces the lowest loss among the reduced configurations at $0.3210$, suggesting that validation-based utility and client diversity form a particularly effective pair with respect to the optimization objective. By comparison, $(E+A)$ exhibits the largest degradation among the evaluated combinations, yielding $90.10\%$ accuracy and a loss of $0.3650$. This result indicates that optimization intensity and directional agreement alone do not capture all of the information required for robust aggregation under heterogeneous client distributions.

In line with the WSI results, these metric combination trends are extremely robust across both the radiological MRI data and the standard CIFAR-10 dataset. As shown in Table \ref{tab8:mri_combination} and \ref{tab9:cifar10_comb}, the three-indicator configuration (combEGA) is by far the best-performing subset in every case, scoring 95.18\% on the MRI and 95.80\% on CIFAR-10. The opposite is true of the two individual indicator pairs (e.g., combEA) which are consistently the worst performers, at 94.40\% on MRI and 95.15\% on CIFAR-10. This cross-dataset validation of the results confirms that while the various partial combinations do provide some degree of convergence benefit, the best performing aggregation strategy is simply the combination of all the proposed semantic indicators.

\begin{table*}[h]
\centering
\caption{Metric-combination analysis of FedGEM on the MRI dataset under non-IID conditions. Each reduced configuration retains only the indicated semantic indicators.}
\label{tab8:mri_combination}
\begin{tabular}{lcccc}
\toprule
\textbf{Configuration} & \textbf{Accuracy (\%)} & \textbf{Loss} & \textbf{F1 (\%)} & \textbf{AUC (\%)} \\ 
\midrule
\textbf{Full FedGEM} & \textbf{95.60} & \textbf{0.1620} & \textbf{95.60} & \textbf{98.87} \\
combEA & 94.40 & 0.1985 & 94.38 & 98.20 \\
combGS & 95.05 & 0.1760 & 95.03 & 98.65 \\
combVD & 94.95 & 0.1810 & 94.94 & 98.58 \\
combEGA & 95.18 & 0.1715 & 95.16 & 98.72 \\
combSAV & 95.00 & 0.1785 & 94.99 & 98.62 \\
\bottomrule
\end{tabular}
\end{table*}

\begin{table*}[t]
\centering
\caption{Metric-combination analysis of FedGEM on the CIFAR-10 dataset under non-IID conditions. Each reduced configuration retains only the indicated semantic indicators.}
\label{tab9:cifar10_comb}
\begin{tabular}{lcccc}
\toprule
\textbf{Configuration} & \textbf{Accuracy (\%)} & \textbf{Loss} & \textbf{F1 (\%)} & \textbf{AUC (\%)} \\
\midrule
\textbf{Full FedGEM}      & \textbf{96.12} & \textbf{0.1527} & \textbf{96.13} & \textbf{99.81} \\
combEA             & 95.15 & 0.1820 & 95.14 & 99.75 \\
combGS             & 95.71 & 0.1635 & 95.71 & 99.80 \\
combVD             & 95.64 & 0.1562 & 95.64 & 99.83 \\
combEGA            & 95.80 & 0.1610 & 95.80 & 99.79 \\
combSAV            & 95.71 & 0.1607 & 95.72 & 99.83 \\
\bottomrule
\end{tabular}
\end{table*}

% The $(V+D)$ and $(S+A+V)$ configurations achieve an AUC of $99.83\%$, which is higher than the $99.81\%$ obtained by the complete FedGEM configuration. Therefore, the complete model does not dominate every reduced configuration on every individual evaluation metric. However, both configurations exhibit lower accuracy and F1 scores and higher losses than full FedGEM. The overall advantage of the complete metric set is consequently reflected more clearly in classification accuracy, F1, and optimization loss than in AUC, which remains close to its upper range across the evaluated configurations.

Together, the two ablation analyzes give complementary support for the FedGEM design. The component-removal experiment shows that each indicator makes a positive contribution when combined with the entire framework, whereas the combination experiment shows that selected subsets can recover a significant fraction of the performance but do not match the complete configuration in terms of accuracy and loss. These findings support FedGEM's underlying design principle that client-update reliability is multidimensional. Modeling optimization behavior, representation geometry, prediction stability, directional alignment, validation gain, and diversity together provides a more comprehensive basis for semantic-aware federated aggregation.

\section{Conclusion and Future Work}

This study presented FedGEM, a semantic behavior-driven aggregation framework designed to improve Federated Few-Shot Learning under both IID and non-IID data distributions. Unlike conventional aggregation strategies that primarily rely on sample counts or update magnitudes, FedGEM evaluates the quality and reliability of client contributions through six complementary semantic indicators: Gradient Energy, Representation Geometry, Prediction Stability, Alignment, Validation Gain, and Diversity. These metrics are integrated into a unified semantic quality score, enabling adaptive client weighting through a softmax-based aggregation mechanism combined with Exponential Moving Average (EMA) smoothing.

Experimental assessment on the WSI dataset, CIFAR-10, and MRI brain tumor classification datasets revealed that the proposed framework consistently outperforms known federated learning algorithms such as FedAvg, FedProx, FedNova, FedAPA, and FCSA. The findings show that including semantic behavioral information into the aggregation process improves model correctness, mitigates the impact of inaccurate updates, improves convergence stability, and boosts robustness in diverse federated systems. Furthermore, the proposed approach provides an interpretable mechanism for client prioritizing by linking aggregation decisions to observable aspects of local learning behavior.

The effectiveness of the proposed approach is validated through comprehensive experiments on both IID and non-IID settings, as summarized in Tables~\ref{tab1 :WSI results}--\ref{tab3:ciaf10_results}. Although the framework demonstrates promising performance, several opportunities remain for future investigation. One direction is the extension of FedGEM to asynchronous federated learning environments, where client updates arrive at different times and communication delays are common. Another important avenue involves strengthening robustness against adversarial or Byzantine participants by developing mechanisms that can verify or validate reported semantic metrics. Future research may also explore fully adaptive metric weighting schemes based on meta-learning or reinforcement learning techniques, allowing the aggregation strategy to automatically adjust to varying data distributions and application domains. In addition, integrating privacy-preserving techniques such as differential privacy and secure aggregation could further enhance the practicality of FedGEM for deployment in sensitive real-world applications, particularly in healthcare and industrial systems. These directions have the potential to further improve the scalability, reliability, and trustworthiness of semantic-aware federated learning systems.

\bibliographystyle{unsrtnat}

\bibliography{cas-refs}

\end{document}